\documentclass[runningheads]{llncs}
\usepackage{makeidx}
\usepackage{graphicx}
\usepackage{amsmath,amssymb} 
\usepackage[caption=false]{subfig}
\usepackage{url}

\newcommand{\ignore}[1]{}

\begin{document}
\pagestyle{headings}
\mainmatter

%
%

%
%
\title{The HAWKwood Database}

%
%
%
\titlerunning{The HAWKwood Database}
\authorrunning{C. Herbon}
  
\author{Christopher Herbon}
\institute{HAWK Fakult\"at Naturwissenschaften und Technik,\\Von-Ossietzky-Stra{\ss}e 99,\\37085 G\"ottingen, Germany\\
\email{herbon@hawk-hhg.de}}

\maketitle

\begin{abstract}
We present a database consisting of wood pile images, which can be used as a benchmark to evaluate the performance of wood pile detection and surveying algorithms. We distinguish six database categories which can be used for different types of algorithms. Images of real and synthetic scenes are provided, which consist of \textbf{7655 images} divided into \textbf{354 data sets}. Depending on the category the data sets either include ground truth data or forestry specific measurements with which algorithms may be compared.
\end{abstract}

\section{Introduction}
The automatic surveying of wood piles has recently drawn attention of computer vision researchers. Over the last decade much works has been done to automatically detect and measure wood piles and individual wood logs (e.g. \cite{Fink2004},  \cite{Gutzeit2010}, \cite{Gutzeit2011}, \cite{Gutzeit2012}, \cite{Herbon2014Detection3D}, \cite{Herbon2014Detection2D}, \cite{Noonpan2013}, \cite{schraml2014temporal}, \cite{schraml2013pith},  \cite{schraml2014similarity})  but the comparison between different methods remains challenging. With our \textit{HAWKwood} database we provide researchers with a large collection of wood pile images and ground truth data, that can be used to evaluate different kinds of wood pile surveying algorithms.

Depending on the country and local conventions different parameters of a wood pile may be of interest. We distinguish between the following three types of parameters which the presented database provides images and ground truth data for.

\begin{enumerate}
	\item The number of wood logs $N$
	\item The solid wood volume $V_s$
	\item The contour volume $V_c$
\end{enumerate}

Furthermore we include some special cases in the \textit{HAWKwood} database, which can help optimizing the detection and segmentation procedures. 

\section{Availability and modifications}

The images in our database are restricted by copyright. The database may be used for non-commercial and educational purposes. A subset of the original images contain sensitive information, that must not be distributed. These images have been manually modified from their original form to remove this information. The database can be downloaded at:\\
\begin{center}
 \textbf{\url{https://193.175.106.27/~cherbon/hawkwood.zip}}
\end{center}

\section{Categories}
In this section we describe the six different subcategories of the benchmark. Each category represents an important aspect of wood log detection and wood pile surveying. We explicitly list the number of data sets, the  over all number of images (real or synthetic), and the ground truth type (manually marked, computed, or determined by forestry standards). The database is divided into two benchmarks, a single and a multi image benchmark, which consist of the following subcategories:\\

\textbf{Single image benchmark}
\begin{itemize}
	\item[ ] Category S.1: wood log detection
	\item[ ] Category S.2: wood log faces segmentation
	\item[ ] Category S.3: front surface segmentation
\end{itemize}

\textbf{Multi image benchmark}
\begin{itemize}
	\item[ ] Category M.1: wood log detection
	\item[ ] Category M.2: solid wood volume computation
	\item[ ] Category M.3: contour volume computation
\end{itemize}

\newpage
\section{Single image benchmark}
\subsection{Category S.1: wood log detection}

\begin{itemize}
\renewcommand{\labelitemi}{$\bullet$}
	\item Number datasets: 36 synthetic, 121 real
	\item Number of images: 36 synthetic, 121 real
	\item Ground truth type: computed (synthetic), manual (real)
\end{itemize}

The images show parts of different wood piles. All images have been selected from the multi image benchmark database, and aim to cover a broad variety of cases, real and synthetic. The position and size of the wood logs have been manually marked by a bounding rectangle and are displayed as a circle. \\

\textbf{Note:} We only consider wood logs that lie completely within the image and that are occluded no more than 25\%. Partially occluded wood logs are marked as such and must not be included in the evaluation.

\begin{figure}
        \centering
                \includegraphics[width=0.2\textwidth]{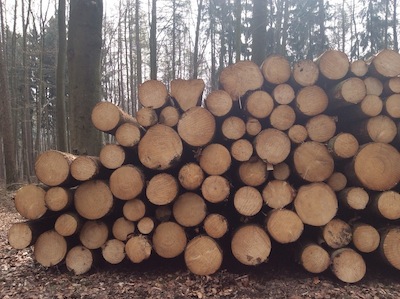}
                \includegraphics[width=0.2\textwidth]{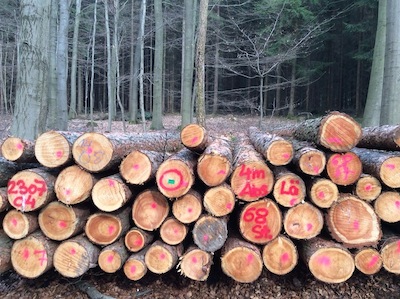}
                \includegraphics[width=0.2\textwidth]{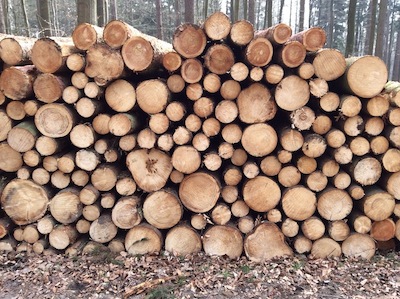}\\
                \includegraphics[width=0.2\textwidth]{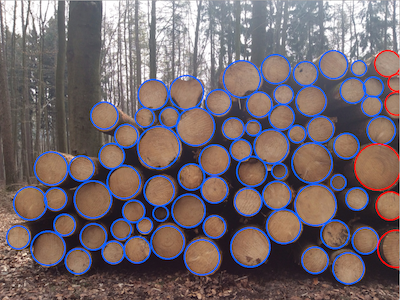}
                \includegraphics[width=0.2\textwidth]{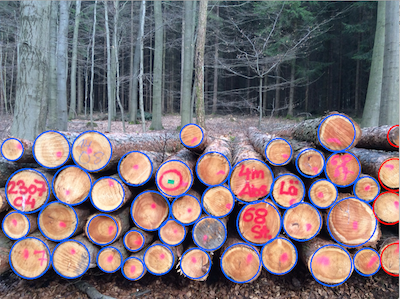}
                \includegraphics[width=0.2\textwidth]{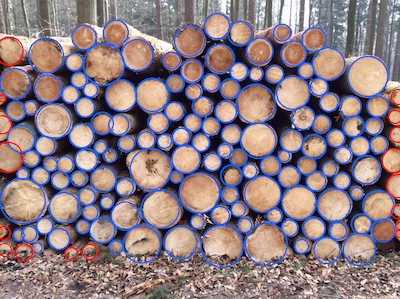}
        \caption{Samples from category S.1. (a)-(c) input images. (d)-(f) manually marked ground truth (blue: valid wood log, red: partially occluded wood log)}\label{fig:samples1.1}
\end{figure}

\subsection{Category S.2: wood log faces segmentation}
\begin{itemize}
\renewcommand{\labelitemi}{$\bullet$}
	\item Number of datasets: 20 (real)
	\item Number of images: 20 images (real) + 20 ground truth masks
	\item Ground truth type: manually marked
\end{itemize}

Category 1.2 includes a subset of the images of category 1.1 where all non occluded wood log front faces have been manually segmented.\\

\textbf{Note:} We only consider wood logs that lie completely within the image and that are occluded no more than 25\%.

\begin{figure}
        \centering
                \includegraphics[width=0.2\textwidth]{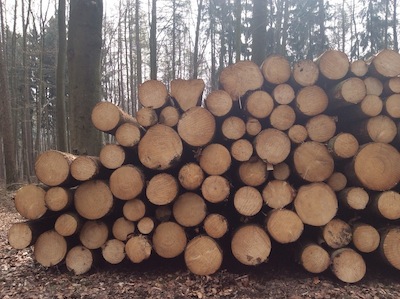}
                \includegraphics[width=0.2\textwidth]{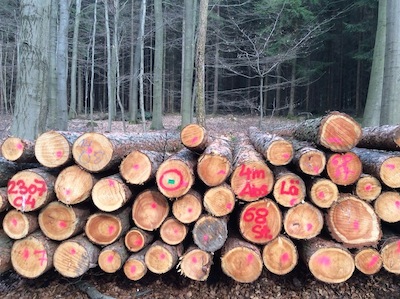}
                \includegraphics[width=0.2\textwidth]{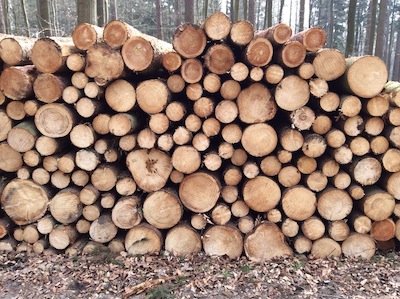}\\
                \includegraphics[width=0.2\textwidth]{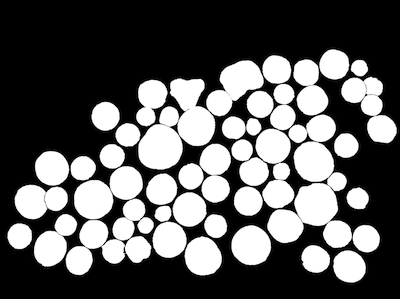}
                \includegraphics[width=0.2\textwidth]{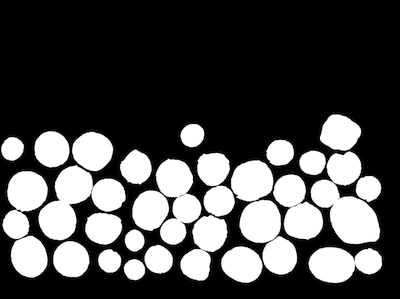}
                \includegraphics[width=0.2\textwidth]{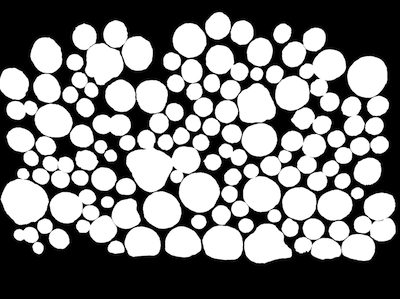}
        \caption{Samples from category S.2. (a)-(c) input images. (d)-(f) binary masks}\label{fig:samples1.2}
\end{figure}

\subsection{Category S.3: front surface segmentation}
\begin{itemize}
\renewcommand{\labelitemi}{$\bullet$}
	\item Number of datasets: 10
	\item Number of images: 10 (real, left) + 10 (real, right) + 10 ground truth masks
	\item Ground truth type: manually segmented
\end{itemize}

\begin{figure}
        \centering
                \includegraphics[width=0.2\textwidth]{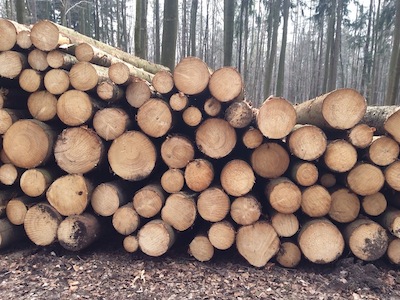}
                \includegraphics[width=0.2\textwidth]{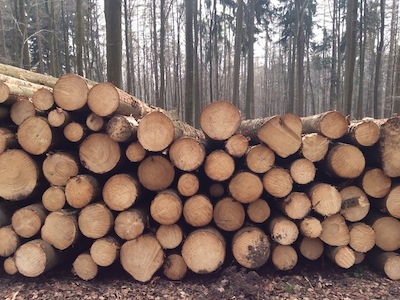}
                \includegraphics[width=0.2\textwidth]{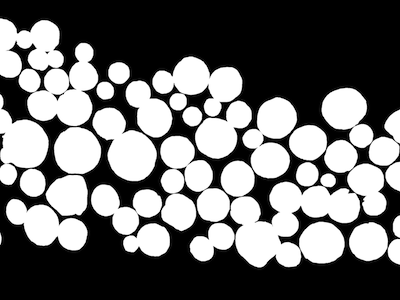}\\
                \includegraphics[width=0.2\textwidth]{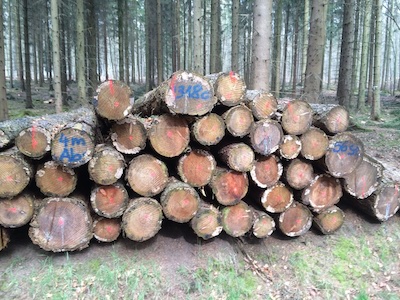}
                \includegraphics[width=0.2\textwidth]{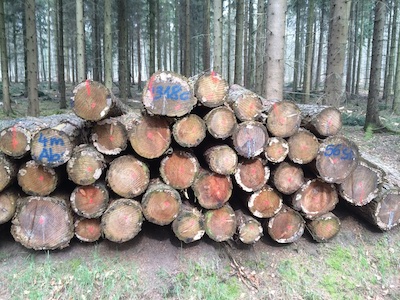}
                \includegraphics[width=0.2\textwidth]{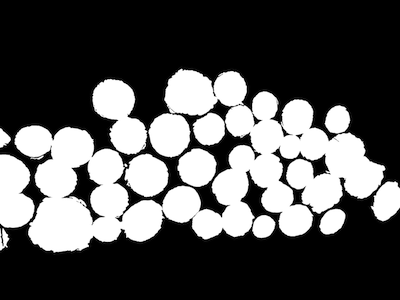}
        \caption{Samples from category S.3. First column: left input image. Second column: right input image. Third column: Manually marked ground truth for the left image.}\label{fig:samples1.3}
\end{figure}

Unlike category 1.2 we consider all visible wood log pixels in category 1.3, even if the wood logs are partially outside of the image area (see figure \ref{fig:samples1.3}). Manually marked, pixel-based ground truth is provided. Furthermore we include a second image, with a slightly translated camera origin. This second image may optionally be used to apply stereo or multiple view algorithms. We explicitly do not provide calibration data, as we consider auto-calibration to be part of the segmentation pipeline.

\newpage
\section{Multi image benchmark}
The multi image benchmark covers the scenario where wood piles are too large to be captured by a single image, as is usually the case. For all multi image datasets we distinguish between \textit{large overlap} and \textit{small overlap} datasets, with regards to two different applications:

\begin{enumerate}
	\item \textit{Large overlap} datasets can be used for multi view reconstruction (structure from motion).
	\item \textit{Small overlap} datasets can be used for panoramic image stitching.
\end{enumerate}

\begin{figure}
        \centering
                \includegraphics[width=0.18\textwidth]{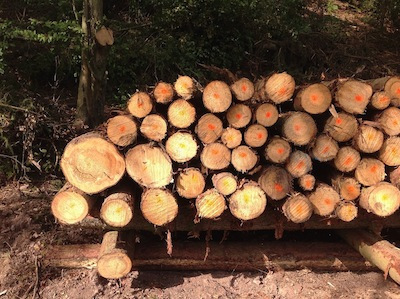}
                \includegraphics[width=0.18\textwidth]{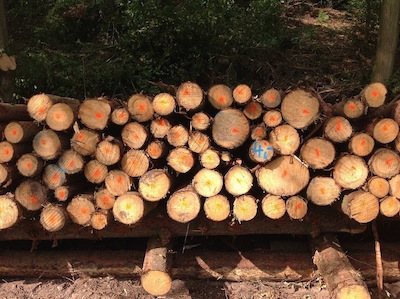}
                \includegraphics[width=0.18\textwidth]{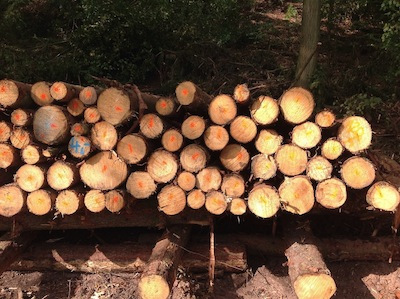}
                \includegraphics[width=0.18\textwidth]{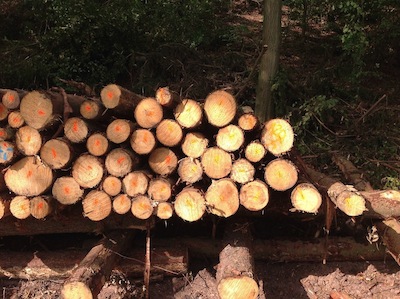}\\
                \includegraphics[width=0.18\textwidth]{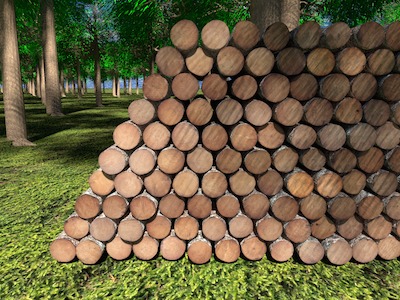}
                \includegraphics[width=0.18\textwidth]{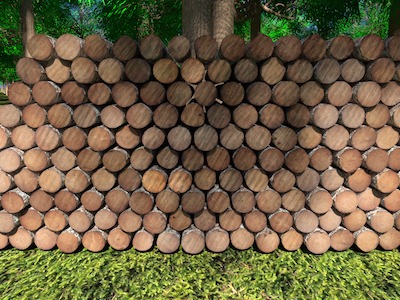}
                \includegraphics[width=0.18\textwidth]{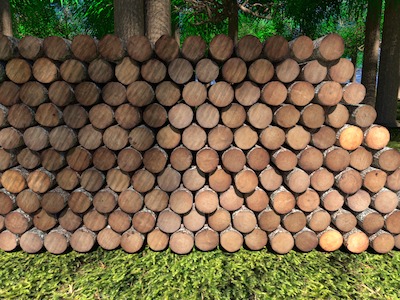}
                \includegraphics[width=0.18\textwidth]{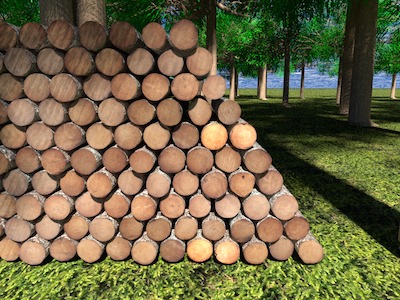}
        \caption{Small overlap samples from category M.1-M.3. Top row real images, bottom row synthetic images.}\label{fig:samples2.1}
\end{figure}

\begin{figure}
        \centering
                \includegraphics[width=0.18\textwidth]{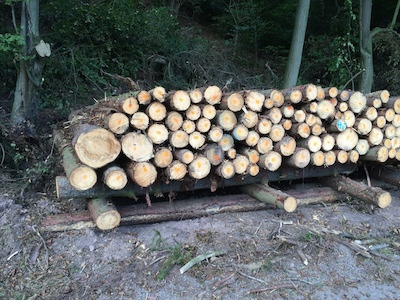}
                \includegraphics[width=0.18\textwidth]{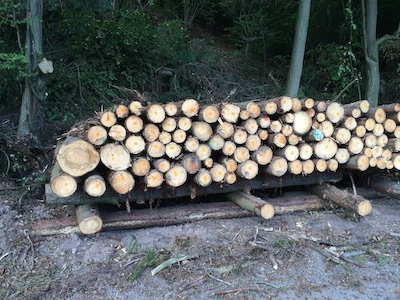}
                \includegraphics[width=0.18\textwidth]{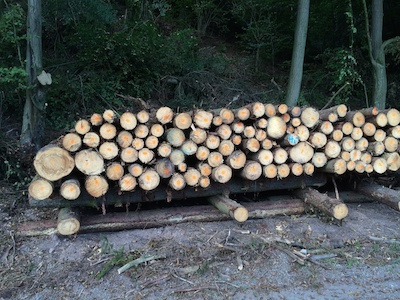}
                \includegraphics[width=0.18\textwidth]{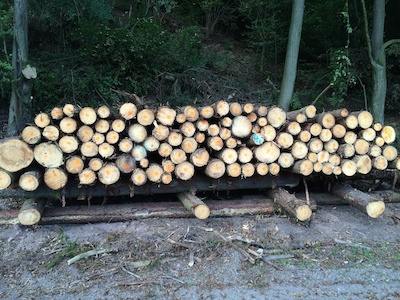}
                \includegraphics[width=0.18\textwidth]{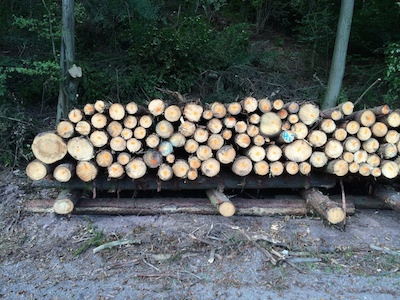}\\
                \includegraphics[width=0.18\textwidth]{cat2/synth/Polter000.jpeg}
                \includegraphics[width=0.18\textwidth]{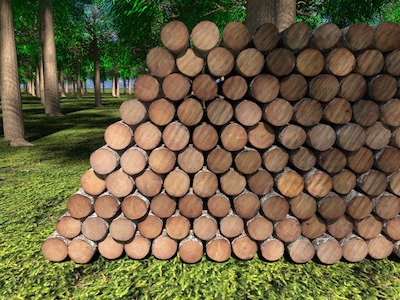}
                \includegraphics[width=0.18\textwidth]{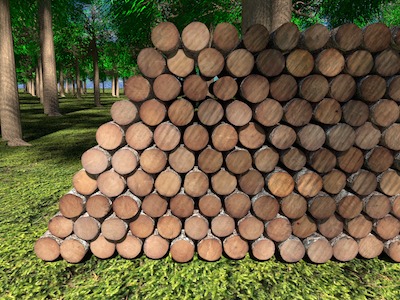}
                \includegraphics[width=0.18\textwidth]{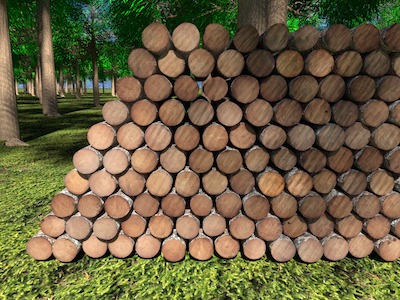}
                \includegraphics[width=0.18\textwidth]{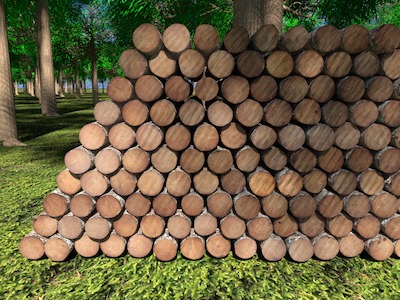}
        \caption{Large overlap samples from category M.1-M.3}\label{fig:samples2.1}
\end{figure}

\subsection{Category M.1: wood log detection}
\begin{itemize}
\renewcommand{\labelitemi}{$\bullet$}
	\item Number of datasets (low overlap): 72  (real)
	\item Number of datasets (high overlap): 147  (real)
	\item Ground truth type: manually counted
\end{itemize}

The number of wood logs has been determined manually. Underlays are per definition not considered part of the actual wood pile and are not included in the ground truth number of logs.

\subsection{Category M.2: solid wood volume computation}
\begin{itemize}
\renewcommand{\labelitemi}{$\bullet$}
	\item Number of datasets (real, low overlap): 34
	\item Number of datasets (real, high overlap): 71
	\item Number of datasets (synthetic, high overlap): 40
	\item Ground truth type:  computed (synthetic), manually measured all wood logs individually (real)
\end{itemize}

The determination of the ground truth solid wood volume is time-consuming and very expensive. It can be done by using a laser scanner and manual post processing. A more cost-efficient approach is using synthetic data, where ground truth can be calculated easily. In this category we provide images of synthetic wood piles as well as real data sets. For the real data sets standard forestry measurements are available. We will refer to the forestry measurements as ground truth approximation.

\begin{figure}[htp]
        \centering
                \includegraphics[width=0.3\textwidth]{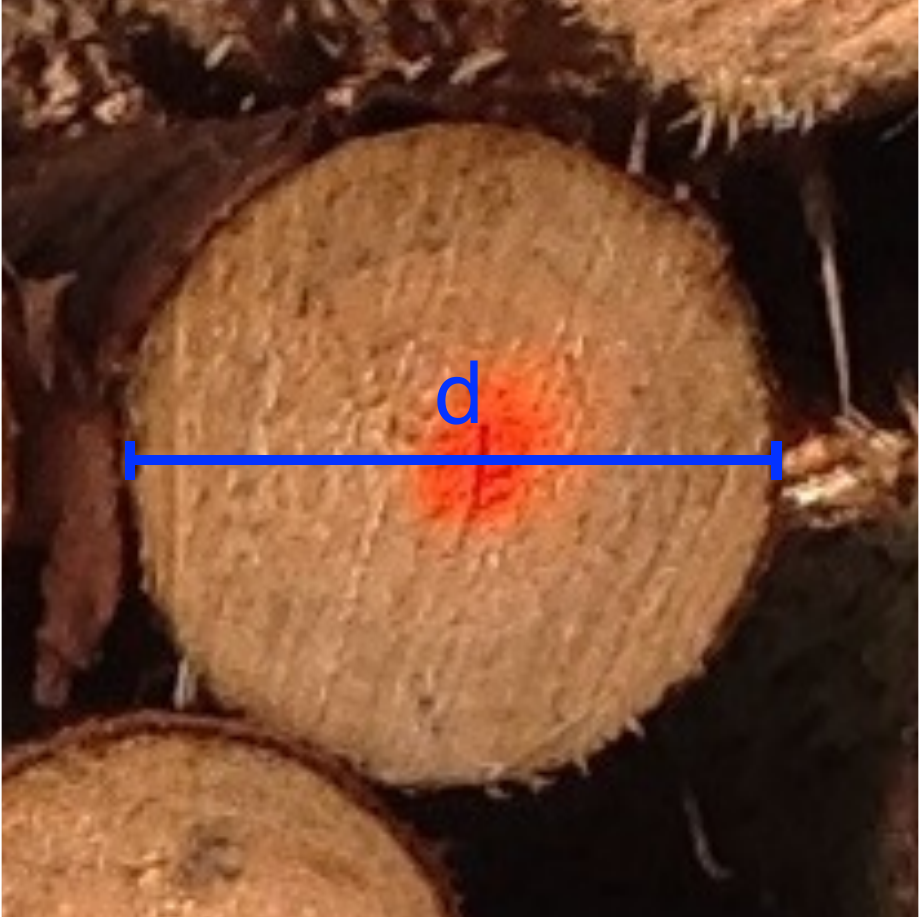}
                \includegraphics[width=0.3\textwidth]{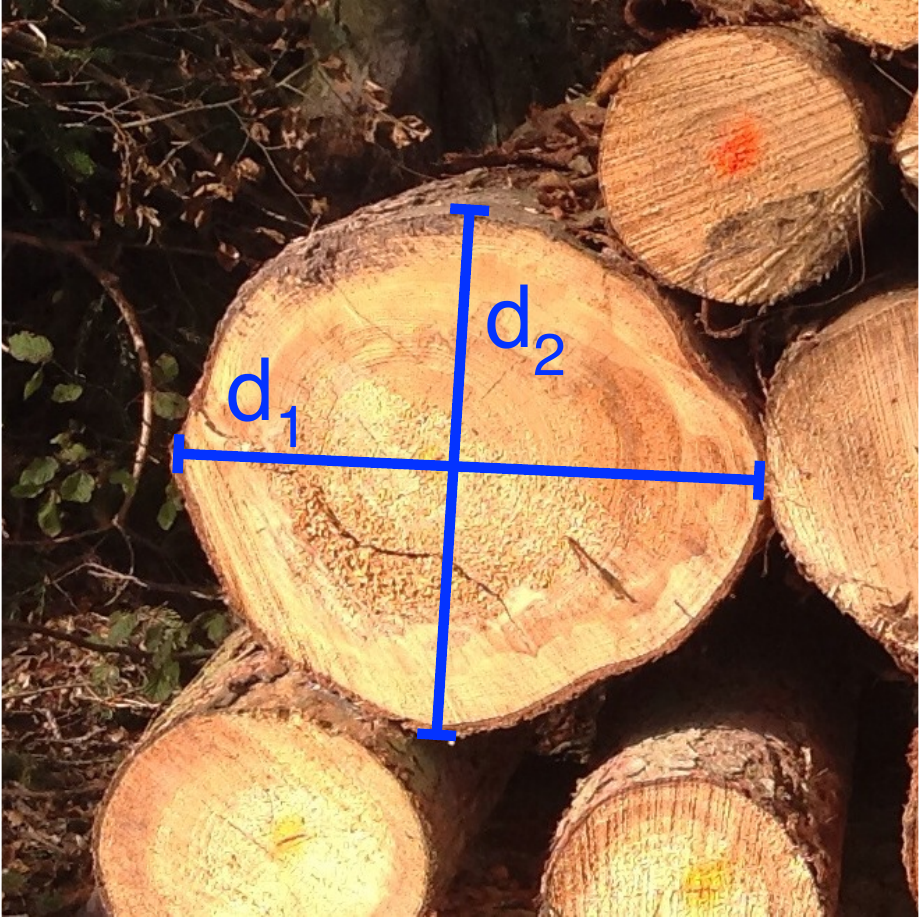}
        \caption{Diameter measurement. (a) Purely horizontal measurement of the diameter for wood logs with $d<20cm$, (b) orthogonal measurement for wood logs with $d\geq 20cm$}
        \label{fig:diameter}
\end{figure}

The ground truth approximation of the solid wood volume is determined as described by Kramer and Ak\c{c}a  \cite{Kramer2008}. If the wood log diameter is smaller than 20cm, the diameter is measured once horizontally. If the diameter is equal to or exceeds 20cm, the diameter of each wood log must be measured twice, with the second diameter being measured orthogonally to the first, while the first diameter should be the largest possible diameter for this wood log. Figure \ref{fig:diameter} illustrates this measurement technique. For all diameter measurements rounding must be performed towards zero \cite{Kramer2008}.

Kramer and Ak\c{c}a \cite{Kramer2008} state that the bark should not be considered in the determination of the wood pile volume. Since there exists different strategies to subtract the bark volume, we must  guarantee comparability despite different bark subtraction methods in different parts of the world. This is achieved by defining the solid wood volume $V_s$ as the sum of all wood log volumes \textit{including} bark, as defined by equation \ref{eq:pilevolume}. We leave it up to each algorithm implementation to subtract the bark volume according to local conventions.

\begin{equation}
\label{eq:pilevolume}
V_s = \sum\limits_{i=1}^N \left(\frac{d_i}{2}\right)^2\pi l
\end{equation}

Equation \ref{eq:pilevolume} uses known and measured quantities, where $N$ is the number of wood logs, $d_i$ is the (average) diameter of each wood log, and $l$ is the known, constant length of each wood log.

Kramer and Ak\c{c}a \cite{Kramer2008} describe different errors, which occur in the measurement pipeline of the wood pile. It is distinguished between systematic and random errors. Random errors emerge from surface irregularities or improper usage of measuring tools. Systematic errors are induced trough flawed measurement equipment and the rounding procedures described above. Since we refer to our measurements as ground truth approximation, these errors are not accounted for.

\subsection{Category M.3: contour volume computation}
\begin{itemize}
\renewcommand{\labelitemi}{$\bullet$}
	\item Number of datasets (real, low overlap): 117
	\item Number of datasets (real, high overlap): 206
	\item Number of datasets (synthetic, high overlap): 40
	\item Ground truth type:  computed (synthetic), forestry standard (real) \cite{Kramer2008}
\end{itemize}

\begin{figure}[htp]
        \centering
                \includegraphics[width=0.7\textwidth]{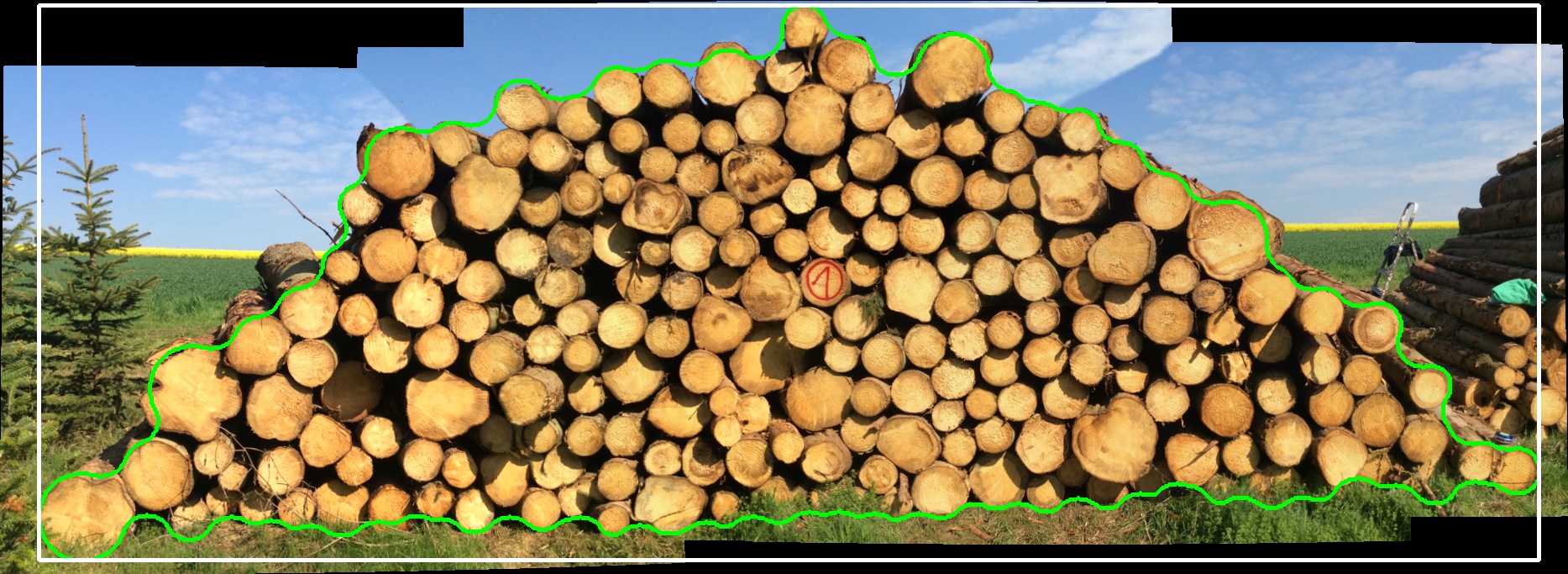}
                \includegraphics[width=0.7\textwidth]{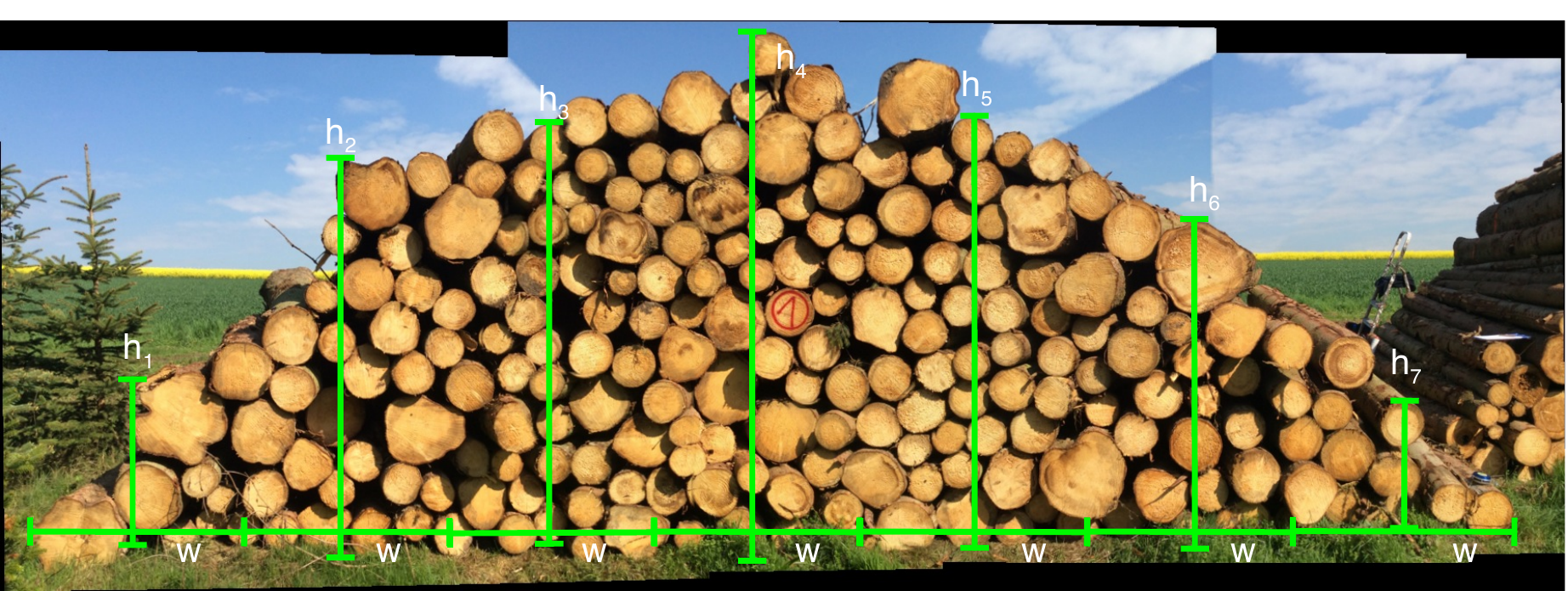}
        \caption{Contour volume measurement. (a) Contour of the wood pile (green), (b) average height computation.}
        \label{fig:contourvolume}
\end{figure}

In many cases the wood quality is poor and the wood is processed in little pieces for e.g. paper production or fire wood. For this type of wood it is not important to know the number of wood logs or the solid wood volume, we are only interested in the volume of the contour (see figure \ref{fig:contourvolume}(a)). We denote the contour volume as $V_c$, which can be calculated by equation \ref{eq:contourvolume}  \cite{RSV88}, where $w_p$ is the width of the wood pile, $l$ is the wood log length, and $h_i$ is the height of the $i$th wood pile section.

\begin{equation}
\label{eq:contourvolume}
V_c = w_p l \frac{1}{k}\sum\limits_{i=1}^k h_i
\end{equation}

Figure \ref{fig:contourvolume} visualizes the calculation of the average height. The wood pile is divided into equal horizontal sections $1\dots k$ and the height of each section $h_i$ is measured in the center is the section.

As we noted before for category M.2, the volume $V_c$ must not be viewed as ground truth but rather as an approximation by forestry conventions. There is no standard as to what width should be used for each section. The convention is to use use a section width of $w$=1.0m. For our data we used $w$=0.5m for more accurate results. We also provide synthetic data sets, where the exact contour volume has been determined.

\section{Future work}
We are constantly looking to extend and improve our database. If you are interested in sharing your images, data, or results please do not hesitate to contact us.

\bibliographystyle{splncs03}
\bibliography{hawkwood}

\end{document}